\documentclass[a4paper]{IEEEtran}
\usepackage{cite}
\usepackage[pdftex]{graphicx}

\ifCLASSINFOpdf
\else
\fi

\usepackage{amsmath}

\usepackage{algorithmic}

\usepackage{dblfloatfix}

\usepackage{url}

\usepackage{multirow}
\usepackage{subfig}

\begin{document}
\title{The Aleatoric Uncertainty Estimation \\ Using a Separate Formulation with Virtual Residuals}

\author{\IEEEauthorblockN{Takumi Kawashima\IEEEauthorrefmark{1},
Qing Yu\IEEEauthorrefmark{1},
Akari Asai\IEEEauthorrefmark{2},
Daiki Ikami\IEEEauthorrefmark{1}\IEEEauthorrefmark{3} and
Kiyoharu Aizawa\IEEEauthorrefmark{1}}
\IEEEauthorblockA{\IEEEauthorrefmark{1}The University of Tokyo\\
 Email: see https://www.hal.t.u-tokyo.ac.jp/lab/en/people.xhtml}
\IEEEauthorblockA{\IEEEauthorrefmark{2}The University of Washington\\
Email: akari@cs.washington.edu}
\IEEEauthorblockA{\IEEEauthorrefmark{3}Nippon Telegraph and Telephone Corporation}}

\maketitle

\begin{abstract}
We propose a new optimization framework for aleatoric uncertainty estimation in regression problems. Existing methods can  quantify the error in the target estimation, but they tend to underestimate it. To obtain the predictive uncertainty inherent in an observation, we propose a new separable formulation for the estimation of a signal and of its uncertainty, avoiding the effect of overfitting. By decoupling target estimation and uncertainty estimation, we also control the balance between signal estimation and uncertainty estimation. We conduct three types of experiments: regression with simulation data, age estimation, and depth estimation. We demonstrate that the proposed method outperforms a state-of-the-art technique for  signal and uncertainty estimation.
\end{abstract}

\IEEEpeerreviewmaketitle

\section{Introduction}
\label{sec:introduction}

There has been rapid progress in deep learning models for a variety of computer vision tasks; such models have increasingly been introduced into real-world applications. It is crucial to quantify the error of model predictions in safety-critical domains such as assisted driving systems and health care. For instance, an autonomous car steering system first processes the sensory inputs, then feeds them to the deep learning models (e.g., depth estimation and semantic segmentation), and finally conducts higher-level decision making based on the models' output ~\cite{bojarski2016end,huval2015empirical}. If the models make nearly random guesses when faced with uncertainty, the result could be a fatal accident. The systems can make reliable high-level decisions if it is given information about the uncertainty. 

Previous studies on Bayesian modeling divided uncertainty into epistemic and aleatoric types~\cite{gal2016uncertainty}. Epistemic uncertainty reflects a lack of data and knowledge: for instance, a model trained on a depth dataset of rural areas might perform poorly on highways, owing to its lack of knowledge about such environments. Epistemic uncertainty can be reduced given sufficient data. By contrast, aleatoric uncertainty is intrinsic to an observation, and adding more data may not reduce it.

\begin{figure}[!t]
\begin{center}
    \includegraphics[width=0.97\columnwidth]{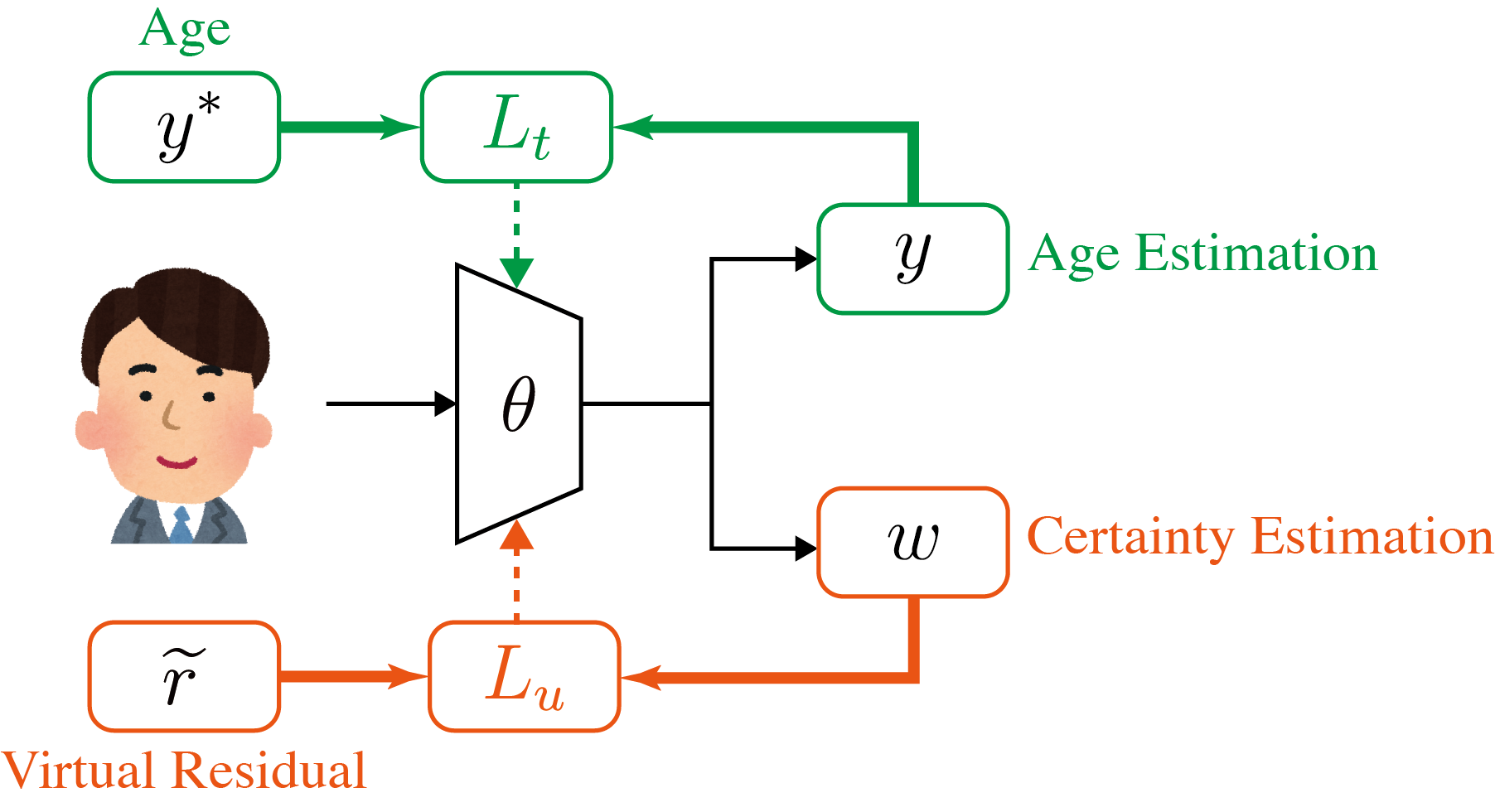}
    \caption{{\bf Outline of the network of  our method}, using the example of age estimation. Here $y$ and $w$ are learned to predict the targets and their certainty, respectively. $L_{t}$ is the loss for the target estimation, and is calculated using the target $y^*$ and its prediction $y$. $L_{u}$ is the loss for the certainty estimation, where we propose using the virtual residual $\tilde{r}_i$ to optimize the certainty estimation $w_i$.}
    \label{fig:introduction}
  \end{center}
\end{figure}

We focus on the aleatoric uncertainty, which cannot be mitigated by increasing the amount of data. Aleatoric uncertainty is further subdivided into homoscedastic uncertainty and heteroscedastic uncertainty. The former is constant for different inputs, while the latter is not. Thus, predicting the heteroscedastic uncertainty from the inputs is particularly important~\cite{NIPS2017_7141}. Unless specified otherwise, the term ``uncertainty'' in this paper refers to heteroscedastic aleatoric uncertainty, i.e., the data-dependent uncertainty from observation noise. 

Previous methods for aleatoric uncertainty estimation have used a Bayesian deep learning framework, in which neural networks predict not only regressed values $y$, but also the uncertainty in the model's prediction~\cite{gal2016uncertainty,NIPS2017_7141}. These methods optimize the estimation of the target and uncertainty at the same time -- the uncertainty estimation is defined based on the target error for the training data. However, usually the error for the test data is far larger than that for the training data, which the model tends to overfit. Thus, in Bayesian formulations, the overfitting of the signal estimation is problematic: when overfitting occurs, the model underestimates errors. There are many ways to mitigate the overfitting, but practically it cannot be removed entirely\cite{overfitting, dropout, stopped}. 

In this paper, we present an optimization framework for aleatoric uncertainty in regression problems by separating the estimation of the target from its uncertainty. In contrast to previous studies, our method allows the network to learn to predict the signal $y$ and certainty $w$ by optimizing the separated loss $L(y, w)=L_t(y)+\lambda L_u(w)$, as shown in Figure~\ref{fig:introduction}. Here, $L_t(y)$ and $L_u(w)$ are the losses of target estimation and certainty estimation respectively, and $\lambda$ is a parameter to balance the weighting of the two terms. Note that the certainty $w$ is estimated instead of the uncertainty itself. We introduce the strict definition of $w$ in Section\ref{sec:related_work}. Our method addresses the shortcomings of existing approaches in two respects. First, we can adjust the two performances of regression and uncertainty estimation by controlling $\lambda$. Second, unlike previous studies of uncertainty estimation, our method explicitly avoids the effect of overfitting. By making use of other models trained with specific samples, we successfully avoid underestimating the error. To demonstrate the effectiveness of our proposed framework, we conduct experiments on three different tasks: simple simulation, age estimation, and depth estimation. Our contributions are as follows:
\begin{itemize}
\item We propose a novel optimization framework for aleatoric uncertainty estimation by a separate formulation, which allows adjusting the weight between the predictive task and the uncertainty estimation, and avoids the effect of overfitting.
    \item We introduce new metrics that measure the goodness of error quantification.
    \item We evaluate our approach on different tasks including age estimation and depth estimation. We find that our method outperforms the state-of-the-art existing methods in terms of error estimation.
\end{itemize}
In our earlier preliminary work, we have proposed a separable formulation without virtual residuals~\cite{asai}. In the last section of this paper, we argue for the benefit of introducing the virtual residuals by comparing our method to the separable formulation without virtual residuals.

\section{Related Work}
\label{sec:related_work}
Although numerous studies have estimated epistemic uncertainty~\cite{gal2016dropout,teye2018bayesian,lakshminarayanan2017simple,Postels2019ICCV}, aleatoric uncertainty has been given less attention. Previous studies have conducted aleatoric uncertainty estimation using Bayesian deep learning frameworks~\cite{gal2016uncertainty,NIPS2017_7141} that train networks to predict data-dependent noise variance from inputs. In Bayesian deep learning, given the input $x$ and the model parameter $\theta$, the network outputs the signal estimation and uncertainty estimation.
The loss function is given by
\begin{equation}
  \mathcal{L}(x, \theta) = \frac{1}{N}\sum_{i=1}^N \left\{ \frac{|y_i^* - y_i(x; \theta)|^2}{2\sigma_i(x, \theta)^2} + \frac{1}{2}\log{\sigma_i(x; \theta)^2} \right\} ,
  \label{eq:kendall_gaussian_original}
\end{equation}
where $y^*$ is the ground truth of $y$. Minimizing this loss function means maximizing the log likelihood of a Gaussian distribution. We call this type of formulation a ``joint formulation'' (JF). Such formulations have previously been used to study applications of heteroscedastic uncertainty ~\cite{bloesch2018codeslam, feng2018leveraging}.

Maximizing the log likelihood of a Laplace distribution rather than the Gaussian distribution of Eq.~\ref{eq:kendall_gaussian_original}, we can use the loss function as below:
\begin{equation}
  \mathcal{L}(x, \theta) = \frac{1}{N}\sum_{i=1}^N \left\{ \frac{|y_i^* - y_i(x; \theta)| }{b_i(x, \theta)} + \log{b_i(x; \theta)} \right\} ,
  \label{eq:kendall_laplace_original}
\end{equation}
Here, $b$ is the scale of the distribution. Kendall et al.~\cite{NIPS2017_7141} used this optimization in their experiments, because it applies $L1$ loss on the target estimation residuals, and seems to outperform the optimization by Eq.~\ref{eq:kendall_gaussian_original}, which applies $L2$ loss.

Eq.~\ref{eq:kendall_laplace_original} can be re-written as Eq.~\ref{eq:kendall_laplace} below, and the network outputs the signal estimation $y$ and the certainty estimation $w$:
\begin{equation}
  \mathcal{L}(x, \theta) = \frac{1}{N}\sum_{i=1}^N \left\{ \exp(w_i(x; \theta)) {r_i(x; \theta)} - w_i(x; \theta)\right\}.
  \label{eq:kendall_laplace}
\end{equation}
In this equation, we introduce two variables: $r_i$ is the absolute value of residual, and $w_i$ is the certainty estimation:
\begin{align}
r_i &= |y_i(x; \theta) - y^*_i|, \label{eq:r_definition} \\
w_i &= -\log b_i. \label{eq:w_definition}
\end{align}


\section{The Influence of Overfitting}
\label{sec:overfitting}
The certainty estimation can be used to estimate the expected value of the error of the target estimation. From the partial derivative of Eq.~\ref{eq:kendall_laplace} at $w_i$, the optimal $w^{opt}_i$ satisfies the following equation:
\begin{equation}
  r_i(x) = \exp(-w^{opt}_i(x)).
  \label{eq:uncertainty_optimization}
\end{equation}
Thus, when we get $y_i$ and $w_i$ as outputs, we can estimate the error of $y_i$ using Eq.~\ref{eq:uncertainty_optimization}. Error estimation plays an important role when we want to know how much we can trust the estimation. However, overfitting hinders using $w_i$ for this purpose.

\begin{figure}[!t]
\begin{center}
    \includegraphics[width=0.75\columnwidth]{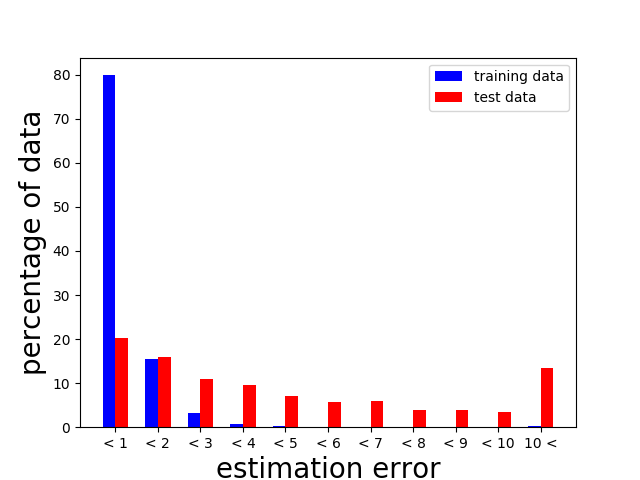}
    \caption{{\bf The distribution of errors in an age estimation task}. The blue and red bars visualize the distribution of the estimation errors of the training and test data, respectively. The estimation was conducted by ResNet50~\cite{he2016deep} and optimized by $L1$ loss. The specifics of the implemaentation are the same as in~\ref{subsec:age_estimation}.}
    \label{fig:error_compare}
  \end{center}
\end{figure}


Fig.~\ref{fig:error_compare} shows the difference of the distribution of the errors between the training and test data for an age estimation task. The estimation errors of the training data are very small -- the majority of them have errors less than one. By contrast, estimation errors are large in the test data. This clearly shows the effect of overfitting, which makes the error of the training data smaller than that of the test data. In its training, we used the validation data, paying careful attention to avoid increasing the error of the validation data. We finish training before the validation error starts ascending. Although we carefully trained the model, the effect of overfitting remained. 

Once overfitting occurs, $r_i$, the error of the target estimation during training, becomes smaller than it should be. Overfitting will make the uncertainty estimation too small if $w_i$ is optimized based on the value of $r_i$ as in the existing method~\cite{NIPS2017_7141}. The aim of our method is to estimate uncertainty suppressing the effect of overfitting.

\section{Proposed Method}
\label{sec:proposed_method}

\subsection{Separate Formulation of Aleatoric Uncertainty}
\label{subsec:separate_formulation}
In this section, we present an optimization framework for uncertainty estimation in regression problems. We focus on training a network that predicts the signal value $y_i$ and its certainty $w_i$.

We cannot balance the weighting of the target estimation and the uncertainty estimation in Eq.~\ref{eq:kendall_laplace} because the two variables for optimization are inseparable. Thus, we propose decoupling the loss function into two terms:
\begin{equation}
  \mathcal{L}(x, \theta) = \frac{1}{N}\sum_{i=1}^N \left\{ L_t(y_i) + \lambda L_u(w_i) \right\}.
  \label{eq:ours_abstract}
\end{equation}
Here, $L_t(y_i)$ is the loss for the target estimation, and $L_u(w_i)$ is that for the uncertainty estimation. By using this formulation, we can balance the target and uncertainty estimation terms by adjusting $\lambda$ appropriately. We define $L_t(y_i) = r_i(x; \theta)$, and $L_u(w_i) = \exp(w_i(x; \theta)) \tilde{r}_i(x; \theta) - w_i(x; \theta)$, resulting in the loss function
\begin{equation}
\begin{split}
& \mathcal{L}(x, \theta) =
\\
& \frac{1}{N}\sum_{i=1}^N \left\{ r_i(x; \theta) + \lambda \left\{ \exp(w_i(x; \theta)) \tilde{r}_i(x; \theta) - w_i(x; \theta) \right\} \right\}.
  \label{eq:ours_laplace}
  \end{split}
\end{equation}
Here, as in Eq.~\ref{eq:kendall_laplace}, we use $L1$ loss and the log likelihood of a Laplace distribution. $\tilde{r}_i$ is a virtual residual which we explain later. The certainty estimation $w_i$ depends on $\tilde{r}_i$ in Eq.9 while it depends on the target residual $r_i$ in Eq.3. By treating the virtual residual $\tilde{r}_i$ as a constant value, $L_t$ works only for the target optimization, and $L_u$ works only for the uncertainty optimization. The virtual residual $\tilde{r}_i$ enables us not only to adjust the balance of the two terms, but also to mitigate the effect of overfitting.


\begin{figure}[!t]
	\begin{center}
  \begin{subfloat}[{\bf}]{
          \includegraphics[clip, width=0.97\columnwidth]{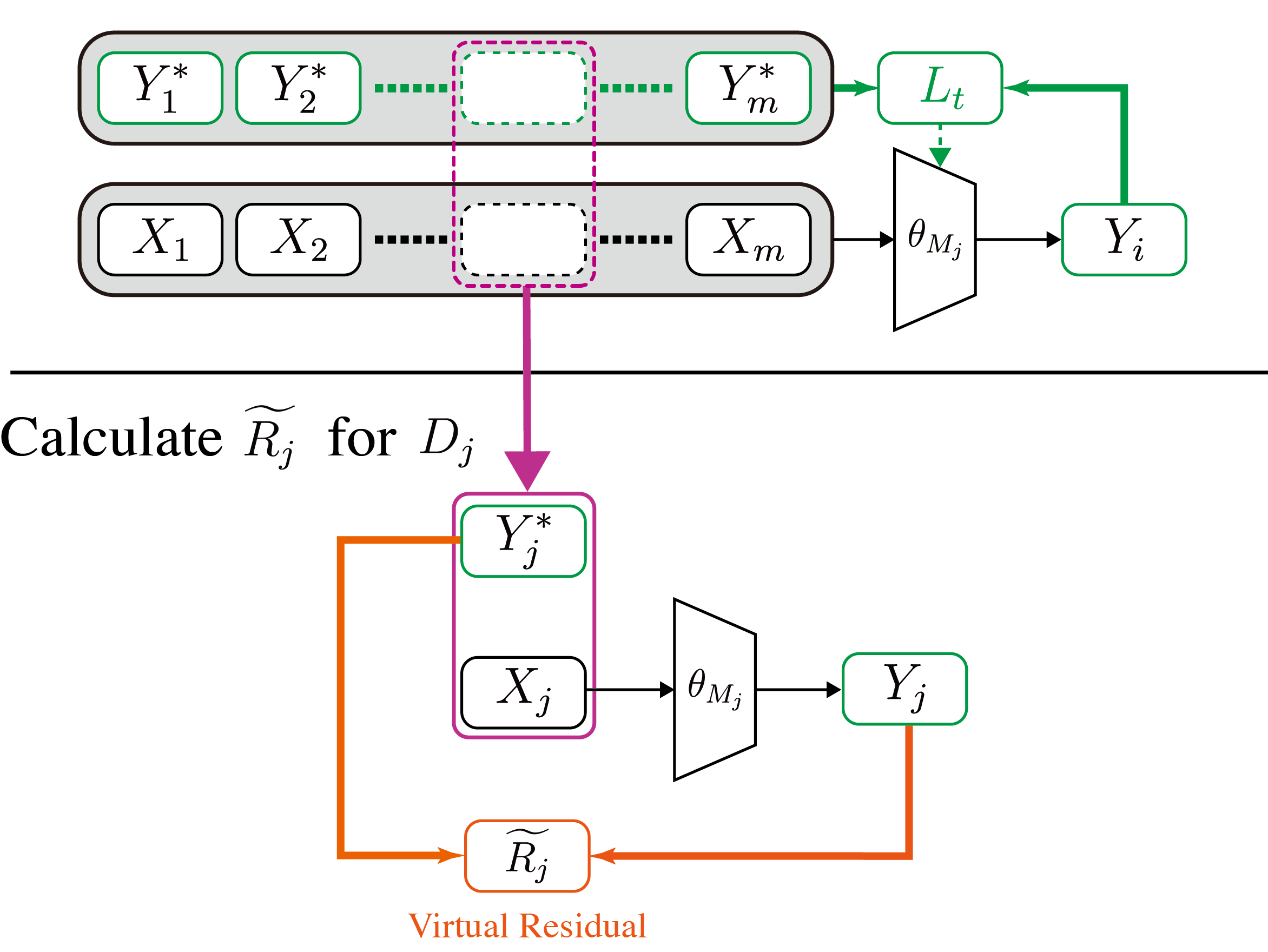}
          \label{fig:proposal_a}}
        \end{subfloat}
        \\
\begin{subfloat}[{\bf}]{
          \includegraphics[clip, width=0.97\columnwidth]{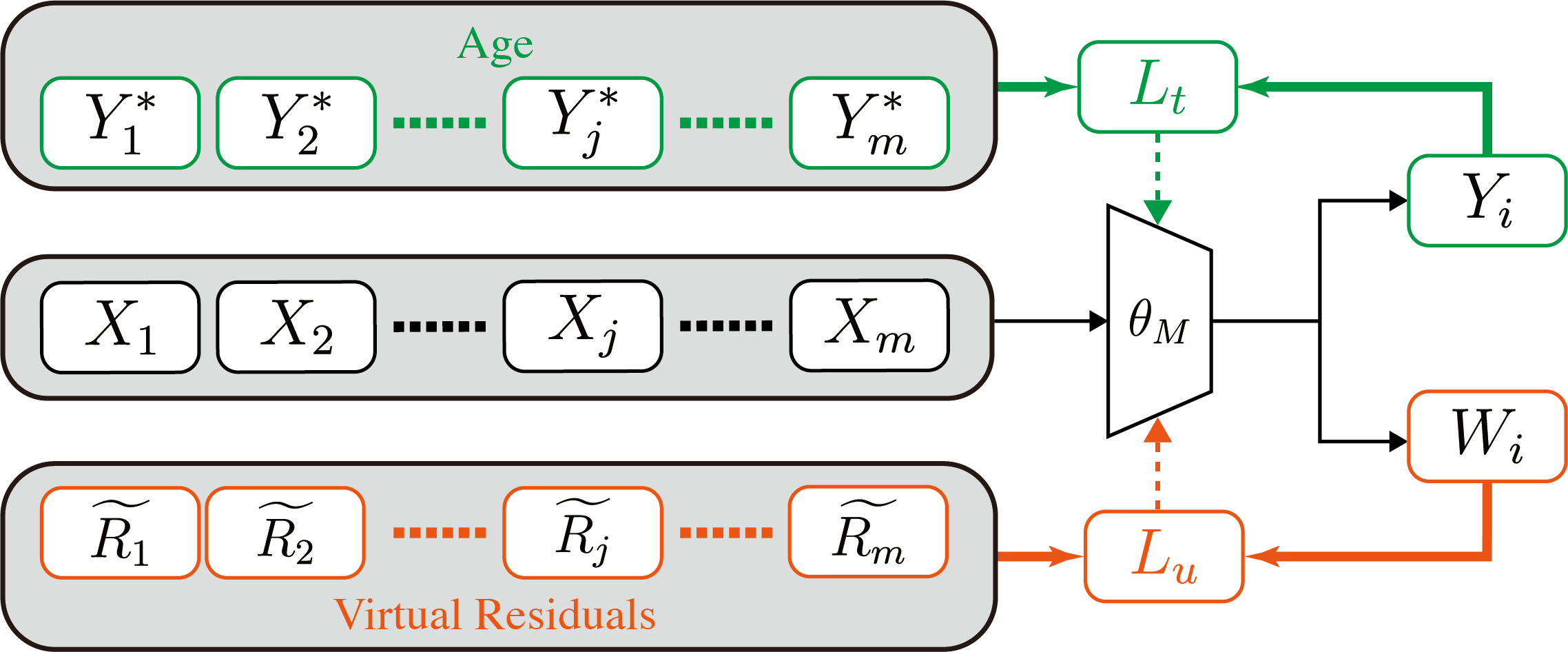}
          \label{fig:proposal_b}}
        \end{subfloat}
	\end{center}

	\caption{{\bf An overview of the use of the virtual residual in our method}, using an example of age estimation. The training data is separated into $m$ subsets $(X_1, Y_1), (X_2, Y_2), ...... , (X_m, Y_m)$. Similarly, we define $m$ models $M_1, M_2, ...... , M_m$. We train each model $M_j$ using the data $\cup\left\{(X_k, Y_k) | k \neq j \right\}$ as shown in the top of Figure~\ref{fig:proposal_a}. $M_j$ is used for calculating the virtual residuals $\tilde{R}_j$ corresponding to the dataset $(X_j, Y_j)$. The virtual residuals $\tilde{R}_1, \tilde{R}_2, ...... , \tilde{R}_m$ are then used when training the model $M$ (which predicts the age and certainty) as shown in Figure~\ref{fig:proposal_b}. 
}
	\label{fig:proposal}
\end{figure}


\subsection{Virtual Residuals}
\label{subsec:virtual_residual}

Next, we discuss the definition of the virtual residuals $\tilde{r}_i$. Fig.~\ref{fig:proposal} illustrates our method. We produce and save the virtual residuals beforehand in the manner described below.

First, in addition to the model $M$ for the target and uncertainty estimation, we define $m$ models $M_1, M_2, ...... , M_m$. Here, $M_j (1 \leq j \leq m)$ has the same construction of layers as $M$, except that $M_j$ outputs only the signal estimation, not the uncertainty estimation. We then randomly separate the training data $D$ into $m$ subsets, namely, $D_1, D_2, ...... , D_m$. Next, we train each model $M_j$ using the data $\cup\left\{D_k | k \neq j \right\}$. Now, we have $m$ trained models, and we let $M_j$ estimate the target values of data subset $D_j$. We then save the error of the signal estimation using the signal estimation from $M_j$. After predicting all $D_j$ by $M_j$, we compute the error $\tilde{r}_i$ for all the training data. Finally, we use $\tilde{r}_i$ in the training by applying Eq.~\ref{eq:ours_laplace}, and train the model $M$ by using all the training data.

We use this virtual residual $\tilde{r}_i$ because it helps us to predict the error, removing the effect of overfitting. When the model estimation of the signal $y_i$ overfits the training data, the residual $r_i$ of training data becomes smaller than that of test data. Therefore if we use Eq.~\ref{eq:kendall_laplace} for training, $w_i$ shows a high certainty, resulting in inaccurate error estimation. In our method, when we optimize the certainty estimation with a sample in $D_j$ for training, we use $\tilde{r}_i$, which is a residual produced from the target prediction of $M_j$. The samples in $D_j$ is not used when we train $M_j$, and thus, the predicted value of $M_j$ is not affected by overfitting to the samples. We can then avoid the certainty estimation being large due to overfitting. Therefore, the certainty estimation $w_i$ of the model $M$ is trained without the effect of overfitting.


\section{Experiments}
\label{sec:experiments}

We applied our method and the existing one to three different tasks. First, we trained the models on toy datasets to demonstrate specific features of the models. Second, we applied the uncertainty estimation to age estimation on the UTKFace dataset. Finally, we estimated uncertainty on an RGB-based depth prediction task, which is inherently difficult. In each of these experiments, we derived the loss function using a Laplacian prior, i.e., Eq.~\ref{eq:kendall_laplace} and Eq.~\ref{eq:ours_laplace}. When $\exp(w_i)$ in Eq.~\ref{eq:kendall_laplace} or Eq.~\ref{eq:ours_laplace} gets too large, we suppress it by replacing the term to $\exp(\frac{w_i}{c})$ with a constant value $c$. We introduce the experimental settings and their results of the three tasks respectively. We find that our method is successful in estimating the error and avoiding underestimation. We also discuss the influence of the virtual residuals by comparing our method with a separable formulation without virtual residuals.


\begin{figure}[!t]
\begin{center}
    \includegraphics[width=0.97\columnwidth]{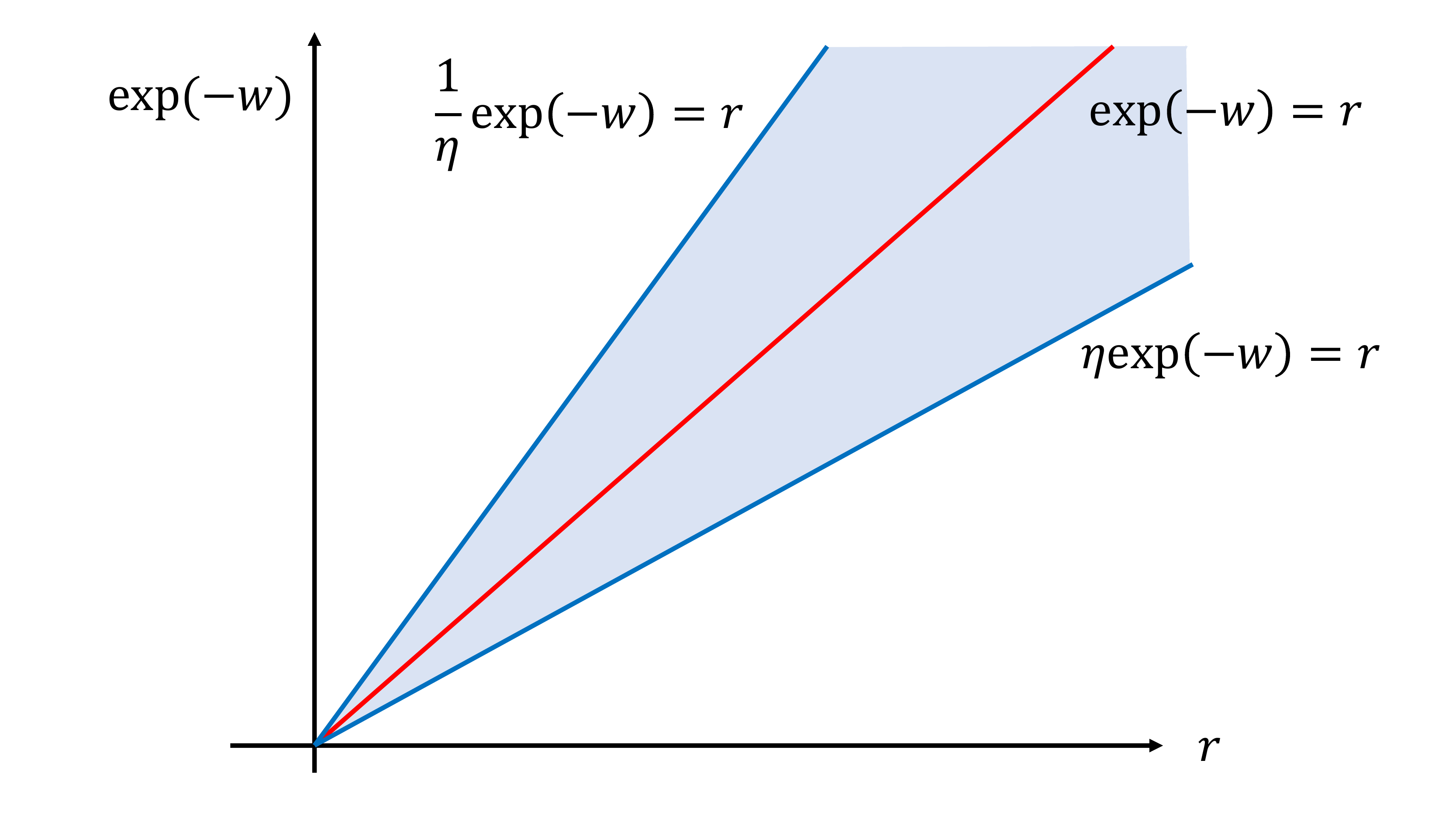}
    \caption{{\bf Visualization of the PiR($\eta$)}. When plotting $(r_i, \exp(-w_i))$ for each sample, PiR($\eta$) shows the rate of the sample that is included in the blue area.}
    \label{fig:pir_definition}
  \end{center}
\end{figure}


\begin{table}[!t]
\small
  \begin{center}
     \caption{{\bf Simulation results}. Showing performance according to metrics (RMSE, eRMSE, and PiR) for simple $L1$ loss estimation, the joint formulation (JF), and our model for various values of $\Delta$ in Eq.~\ref{eq:simulation_function_right}.}
      \begin{tabular}{|c|c||c||c|c|} \hline
    $\Delta$ & Loss & Target & eRMSE & PiR ($0.5$) \\ 
    & & RMSE ($\downarrow$) & ($\downarrow$) & ($\uparrow$) \\ \hline \hline
  \multirow{4}{*}{$0$} 
  & $L1$ & 1.39 & - & -  \\ \cline{2-5}
  & JF~\cite{NIPS2017_7141} & 1.40 & 0.905 & 0.498 \\ \cline{2-5}
  & our method ($\lambda = 0.1$) & 1.40 & {\bf 0.857} & {\bf 0.554} \\ \hline
 \multirow{4}{*}{$1$} 
  & $L1$ & 1.42 & - & - \\ \cline{2-5}
  & JF & 1.39 & 0.890 & 0.504 \\ \cline{2-5}
  & our method ($\lambda = 0.02$) & 1.42  & {\bf 0.858} & {\bf 0.550} \\ \hline
 \multirow{4}{*}{$5$} 
  & $L1$ & 1.40 & - & - \\ \cline{2-5}
  & JF & 1.43 & 0.930 & 0.474 \\ \cline{2-5}
  & our method ($\lambda = 0.1$) & 1.41 & {\bf 0.853} & {\bf 0.566} \\ \hline
  \end{tabular}
\label{tbl:simulation_results}
  \end{center}
\end{table}


\begin{figure}[!t]
	\begin{center}
  \begin{subfloat}[{\bf $\Delta = 0$.}]{
          \includegraphics[clip, width=0.47\columnwidth]{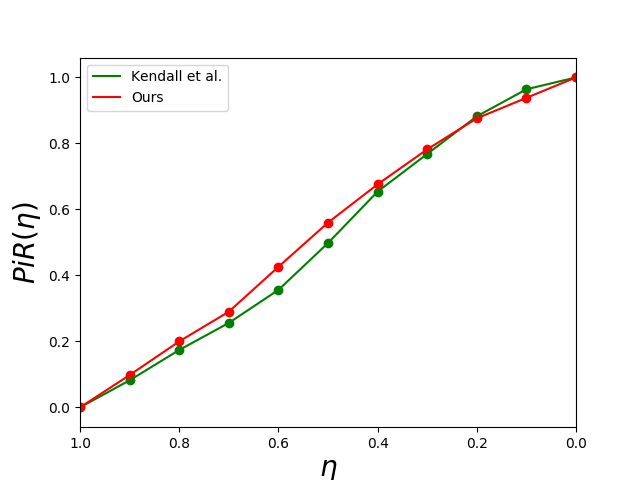}}
        \end{subfloat}
\begin{subfloat}[{\bf $\Delta = 1$.}]{
          \includegraphics[clip, width=0.47\columnwidth]{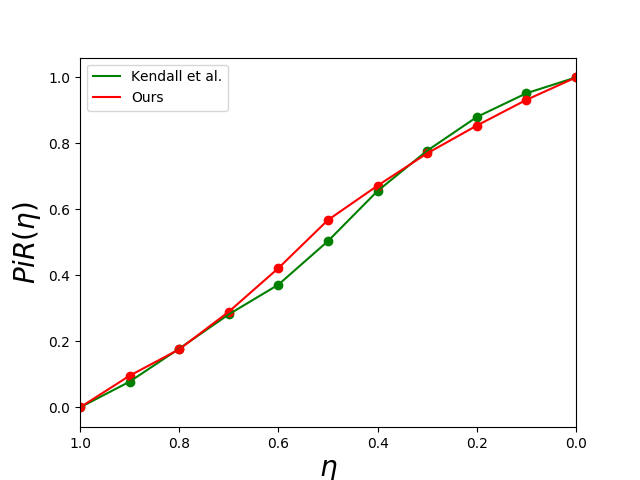}}
        \end{subfloat}
\begin{subfloat}[{\bf $\Delta = 5$.}]{
          \includegraphics[clip, width=0.47\columnwidth]{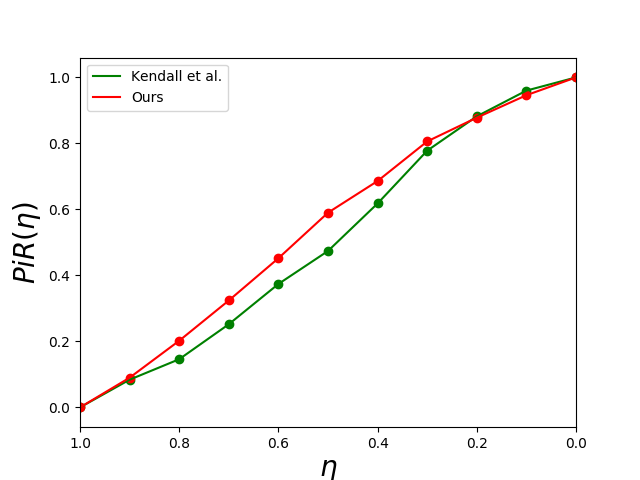}}
        \end{subfloat}
	\end{center}

	\caption{{\bf PiR of the simulation.} 
Each image shows the value of PiR($\eta$) as $\eta$ decreases by $0.1$ from $1$ to $0$. The green and red curves are the results of the joint formulation ~\cite{NIPS2017_7141} and our method, respectively.
}
	\label{fig:simulation_pir}
\end{figure}


\begin{figure*}[!ht]
	\begin{center}
  \begin{subfloat}[{The signal estimatino with JF while $\Delta = 0$.}]{
          \includegraphics[clip, width=0.31\columnwidth]{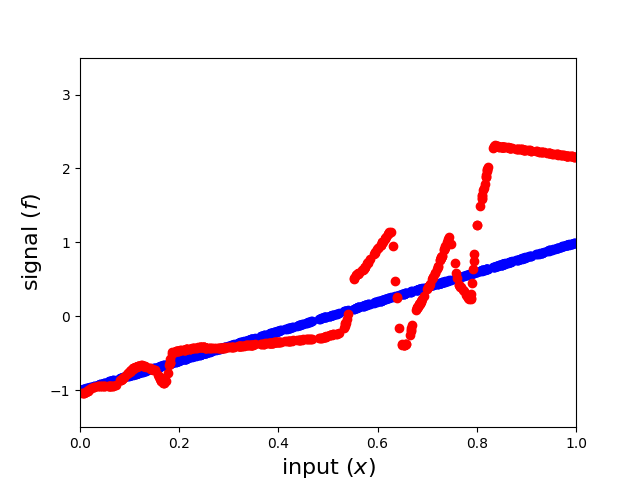}}
        \end{subfloat}
\begin{subfloat}[{The signal estimation with JF while $\Delta = 1$.}]{
          \includegraphics[clip, width=0.31\columnwidth]{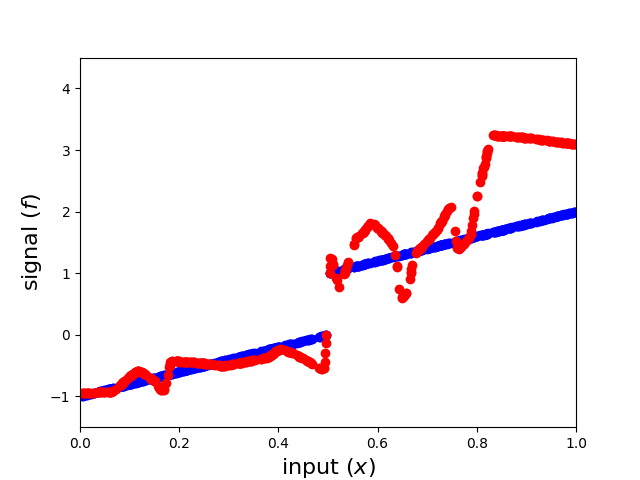}}
        \end{subfloat}
\begin{subfloat}[{The signal estimation with JF while $\Delta = 5$.}]{
          \includegraphics[clip, width=0.31\columnwidth]{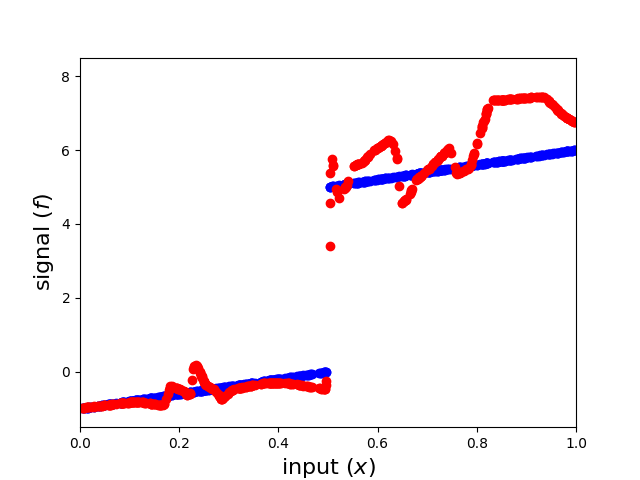}}
        \end{subfloat}
        \begin{subfloat}[{The error estimation with JF while $\Delta = 0$.}]{
          \includegraphics[clip, width=0.31\columnwidth]{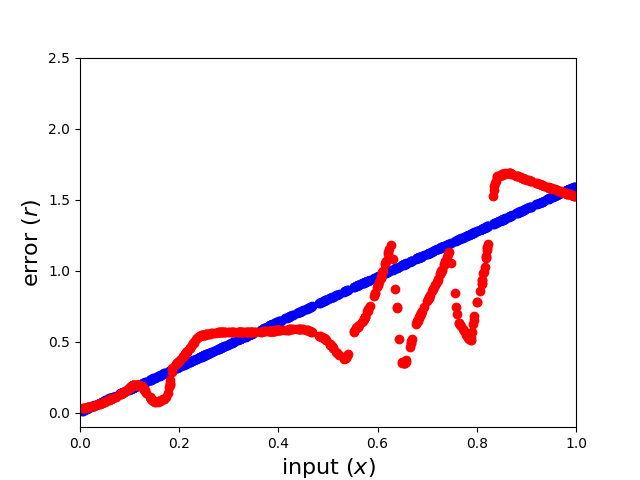}}
        \end{subfloat}
\begin{subfloat}[{The error estimation with JF while $\Delta = 1$.}]{
          \includegraphics[clip, width=0.31\columnwidth]{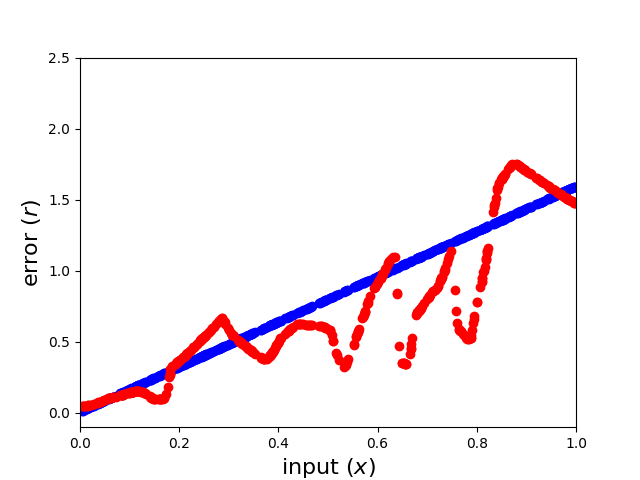}}
        \end{subfloat}
\begin{subfloat}[{The error estimation with JF while $\Delta = 5$.}]{
          \includegraphics[clip, width=0.31\columnwidth]{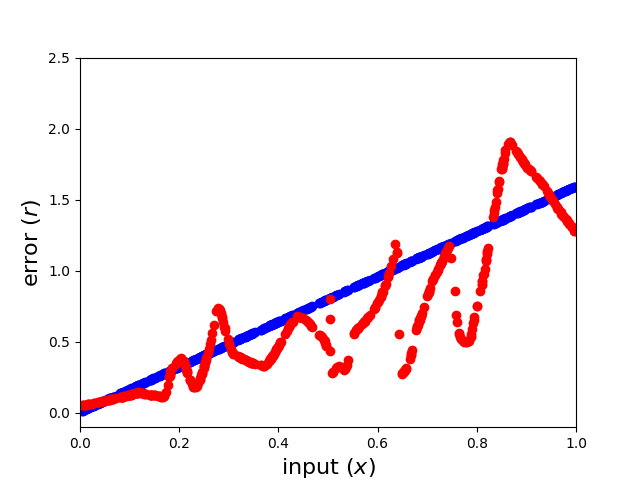}}
        \end{subfloat}
         \begin{subfloat}[The signal estimation with our method while $\Delta = 0$.]{
        \includegraphics[clip, width=0.31\columnwidth]{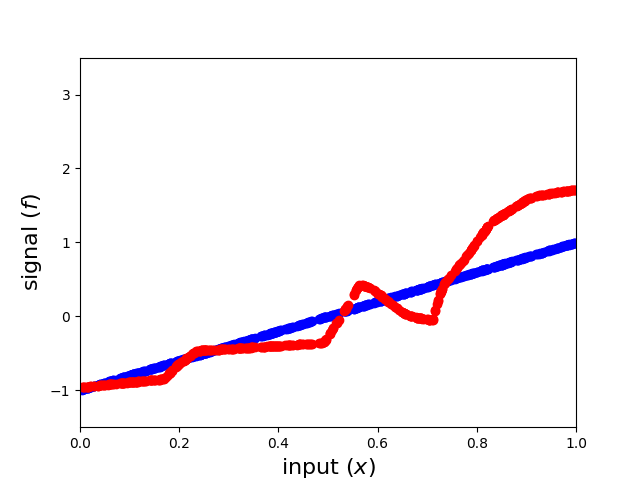}}
        \end{subfloat}
\begin{subfloat}[{The signal estimation with our method while $\Delta = 1$.}]{
          \includegraphics[clip, width=0.31\columnwidth]{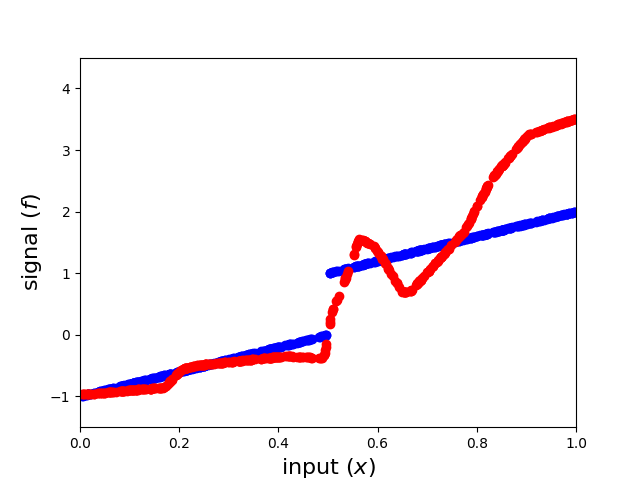}}
        \end{subfloat}
\begin{subfloat}[{The signal estimation with our method while $\Delta = 5$.}]{
          \includegraphics[clip, width=0.31\columnwidth]{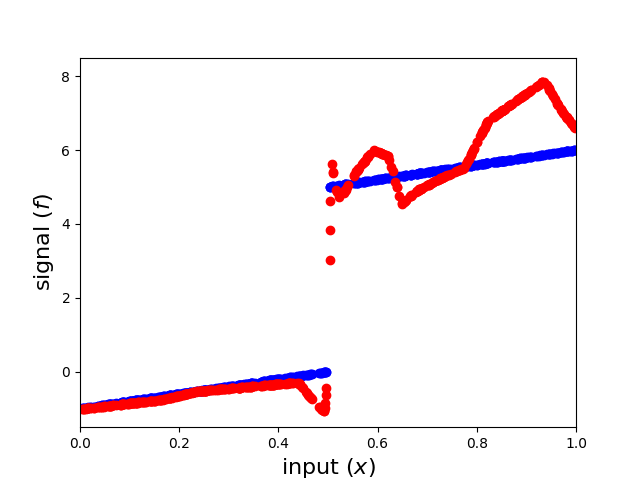}}
        \end{subfloat}
        \begin{subfloat}[{The error estimation with our method while $\Delta = 0$.}]{
          \includegraphics[clip, width=0.31\columnwidth]{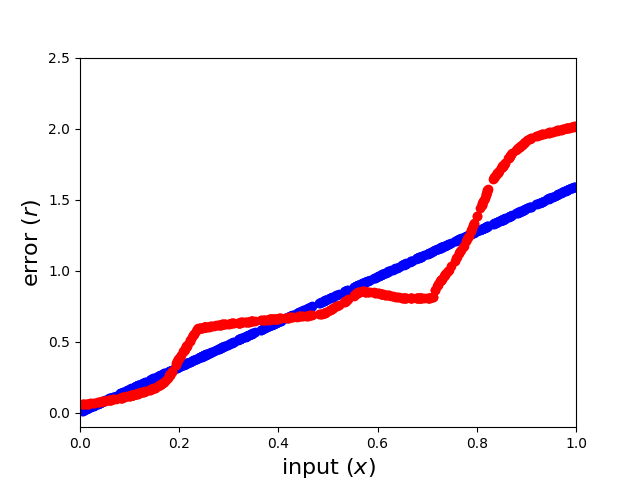}}
        \end{subfloat}
\begin{subfloat}[{The error estimation with our method while $\Delta = 1$.}]{
          \includegraphics[clip, width=0.31\columnwidth]{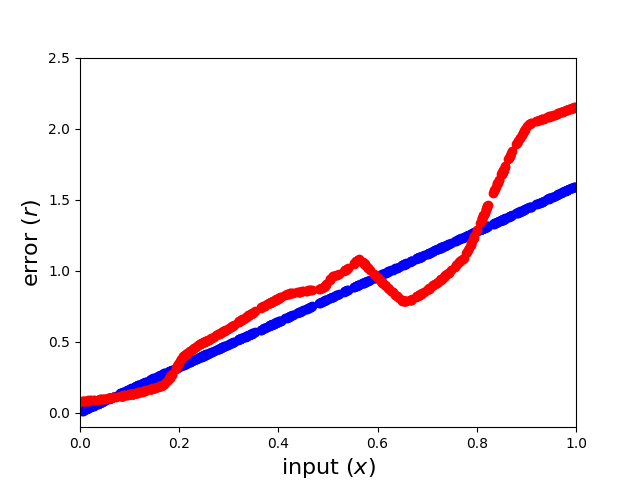}}
        \end{subfloat}
\begin{subfloat}[{The error estimation with our method while $\Delta = 5$.}]{
          \includegraphics[clip, width=0.31\columnwidth]{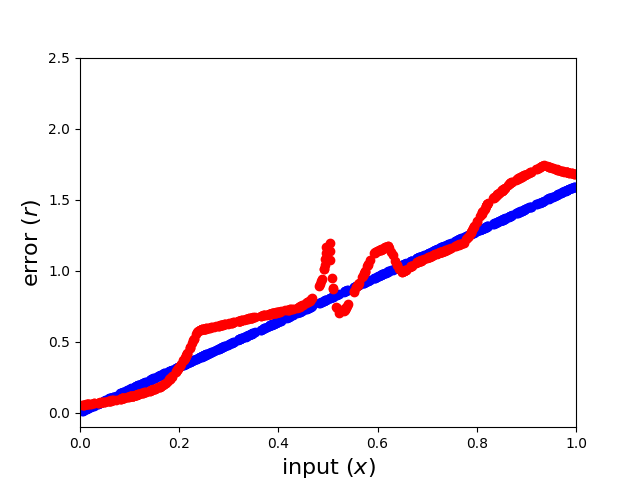}}
        \end{subfloat}
	\end{center}
  \caption{{\bf Signal and uncertainty predictions during the simulation.}  The upper figures show the joint formulation (JF) results~\cite{NIPS2017_7141}, while the lower figures show those of our method. In the six figures on the left side ((a), (b), (c), (g), (h), and (i)), the $x$-axis corresponds to the input, and the $y$-axis to the signal. Blue points indicate the training data; red points are the models' predictions for the test data. In the other six figures on the right side, the $x$-axis corresponds to the input, and the $y$-axis corresponds to the error of the signal estimation. Blue points indicate the ideal mean error given by $\sigma(x)$, and red points are estimations of the error by the models calculated by Eq.~\ref{eq:uncertainty_optimization}. Each point corresponds to one test datum. }
  \label{fig:simulation_visualization}
\end{figure*}


\subsection{Metrics for Uncertainty Estimation}
\label{subsec:metrics}

First of all, we introduce the metrics that we use in our experiments. Sparsification plots are widely used for accurate evaluation of the performance of uncertainty estimation~\cite{wannenwetsch2017probflow,bruhn2006confidence,mac2012learning,eddy2018opticalflow}. In these plots, samples are removed in order of the uncertainty. The plots themselves do not show the uncertainty estimation performance numerically. In this work, we focus on how accurately the model can estimate the value of the error. We evaluate the performance of the error estimation using the following two metrics: 
\begin{itemize}
    \item {\bf eRMSE}: To evaluate the goodness of the certainty value $w_i$ based on Eq.~\ref{eq:uncertainty_optimization},
we introduce a metric eRMSE defined as:
\begin{equation}
  eRMSE = \sqrt{E\left[| \exp(-w_i(x)) - r_i(x) |^2\right]}.
  \label{eq:ermse_definition}
\end{equation}
This metric indicates how well the uncertainty is estimated.
\item {\bf PiR}: We check whether each sample satisfies
\begin{equation}
  \eta \exp(-w_i(x)) \leq r_i(x) < \frac{1}{\eta} \exp(-w_i(x)),
  \label{eq:pir_definition}
\end{equation}
where $\eta$ is a fixed value such that $0 < \eta \leq 1$. We define the rate of the samples satisfying Eq.~\ref{eq:pir_definition} as PiR($\eta$). Considering Eq.~\ref{eq:uncertainty_optimization}, when we plot a point $(r_i, \exp(-w_i))$ for each sample on a two-dimensional plane, ideally the points are located near the line $\exp(-w) = r$. PiR($\eta$) indicates the rate of the points that is located between two lines $\frac{1}{\eta} \exp(-w) = r$ and $\eta \exp(-w) = r$ as Figure~\ref{fig:pir_definition} shows, and thus, it works as another metric indicating how well the uncertainty is estimated.
\end{itemize}
In addition, we compare the performance of target estimation using the root mean square error (RMSE).


\begin{table}[!t]
\small
  \begin{center}
      \caption{{\bf Age estimation results on the UTKFace dataset}.}
\label{tbl:age_results}
      \begin{tabular}{|c||c||c|c|} \hline
    Loss & Target & eRMSE & PiR($\eta$) \\  
    & RMSE ($\downarrow$) & ($\downarrow$) & ($\uparrow$) \\ \hline \hline
    $L1$  & 7.37 & - & - \\ \hline
    JF~\cite{NIPS2017_7141} & 7.51 & 6.10 & 0.352 \\ \hline
    our method ($\lambda = 0.02$) & 7.27 & {\bf 4.99} & {\bf 0.523} \\ \hline
  \end{tabular}
  \end{center}
\end{table}


\begin{figure}[!t]
\begin{center}
    \includegraphics[width=0.5\columnwidth]{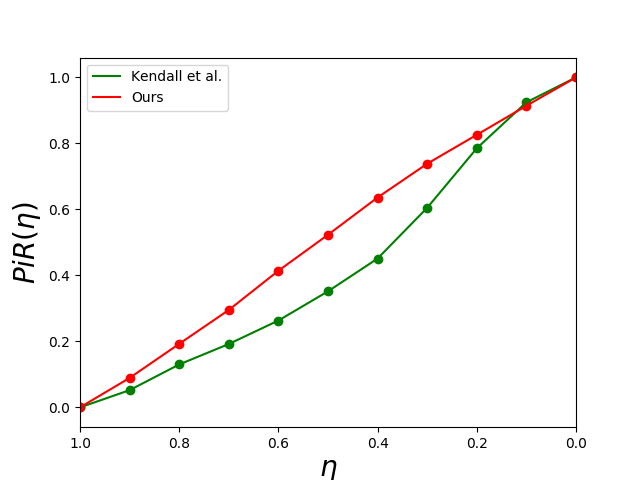}
    \caption{{\bf PiR of the age estimation.} 
This plot shows the value of PiR($\eta$), where $\eta$ decreases by $0.1$ from $1$ to $0$. The green and red line illustrate the results of the joint formulation~\cite{NIPS2017_7141} and our method, respectively.
}
    \label{fig:age_pir}
  \end{center}
\end{figure}


\subsection{Simulation}
\label{subsec:simulation}

In this experiment, we perform a one-dimensional regression task to clearly visualize the behavior of the signal and uncertainty estimations.

\paragraph{Datasets}

We randomly generate input $\{x | 0 \leq x < 1\}$ from a uniform distribution.
The target $y(x)$ is generated from the signal $f(x)$ and noise $n(x)$ as below:
\begin{equation}
  y(x) = f(x) + n(x).
  \label{eq:simulation_target_definition}
\end{equation}
$f(x)$ is defined as
\begin{align}
  f(x) &= 2x - 1 (0 \leq x < 0.5), \label{eq:simulation_function_left} \\
  f(x) &= 2x - 1 + \Delta (0.5 \leq x < 1).  \label{eq:simulation_function_right}
\end{align}
We choose three different $\Delta$ values: $0$, $1$, and $5$.
Note that $f(x)$ is a linear function without a step when we choose $0$ for $\Delta$.


\begin{table}[!t]
\begin{center}
  \caption{{\bf Examples of age and error estimation by two methods (joint formulation and ours)}.}
  \label{tbl:age_examples}
  \begin{tabular}{|c|c||c||c|c|c|}
    \hline input & age & Loss & age & prediction & error \\ 
    & & & estimation & error & estimation \\ \hline
    \hline
  \multirow{2}{*}{
\begin{minipage}{8mm}
      \centering
      \scalebox{0.08}{\includegraphics{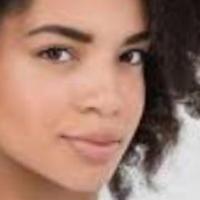}}
    \end{minipage}
  }
  &
  \multirow{2}{*}{20}
	& JF~\cite{NIPS2017_7141} & 23.131 & 3.131 & 1.499 \\ \cline{3-6}
  & & our method & 22.692 & 2.692 & 2.655 \\ \hline \hline
  \multirow{2}{*}{
\begin{minipage}{8mm}
      \centering
      \scalebox{0.08}{\includegraphics{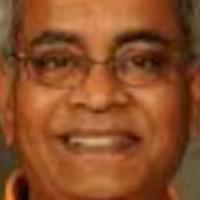}}
    \end{minipage}
  }
  &
  \multirow{2}{*}{68}
	& JF & 59.401 & 8.599 & 2.487 \\ \cline{3-6}
  & & our method & 58.005 & 9.995 & 6.553 \\ \hline \hline
  \end{tabular}
  \end{center}
\end{table}


Noise $n(x)$ is dependent on the input and generated by the Gaussian function $\mathcal{N}(0, \sigma(x)^2)$.
For $\sigma(x)$, we choose a linear function that has a large value when $x$ is large:
\begin{equation}
  \sigma(x) = 1.99x + 0.01.
  \label{eq:simulation_variance}
\end{equation}
In our experiments, given a noisy target $y_i(x)$, we estimate the signal $f_i(x)$ and certainty $w_i(x)$; the latter is taken to be $-\log(|n_i(x)|)$.

We generate 128 samples for training and 500 samples for testing. We train models with relatively small datasets to simulate a finite dataset for large multidimensional tasks and use more test samples for better visualization of the results.


\begin{figure}[!t]
	\begin{center}
  \begin{subfloat}[The age estimation with JF.]{
          \includegraphics[clip, width=0.47\columnwidth]{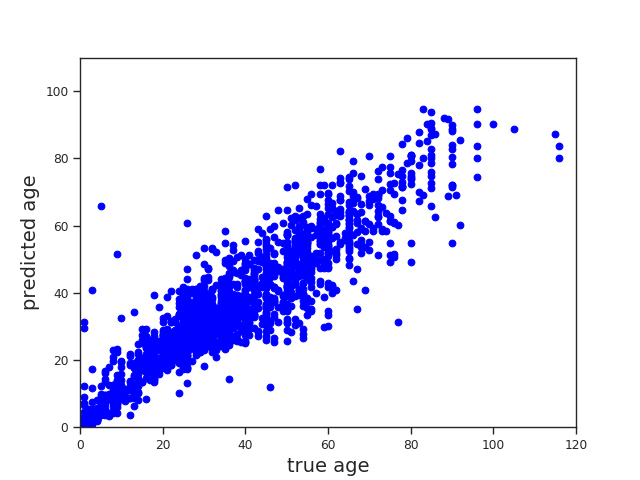}}
        \end{subfloat}
\begin{subfloat}[The error estimation with JF.]{
          \includegraphics[clip, width=0.47\columnwidth]{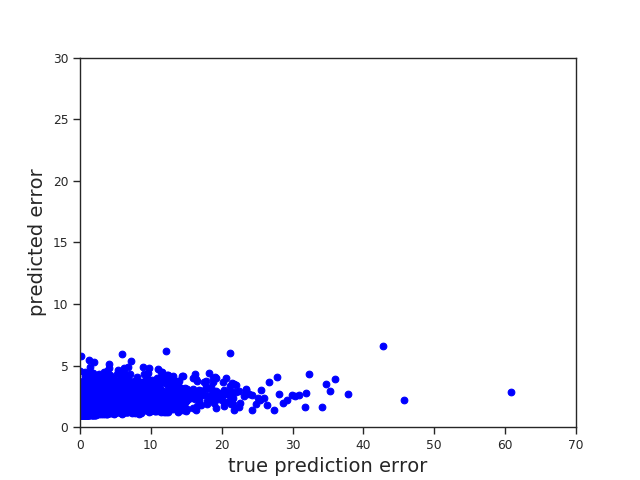}}
        \end{subfloat}
\begin{subfloat}[The age estimation with our method.]{
          \includegraphics[clip, width=0.47\columnwidth]{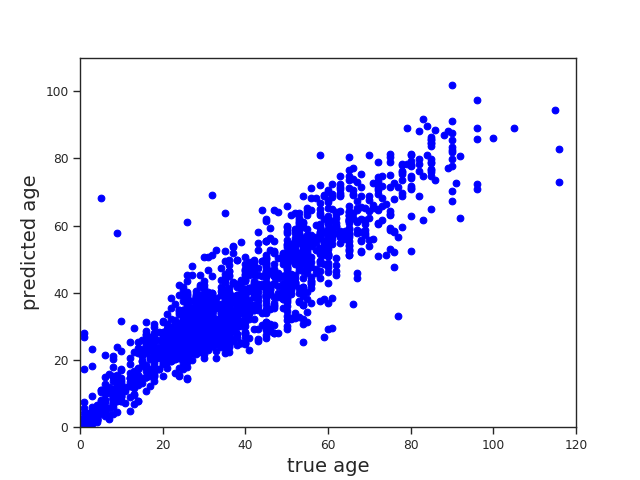}}
        \end{subfloat}
        \begin{subfloat}[The error estimation with our method.]{
          \includegraphics[clip, width=0.47\columnwidth]{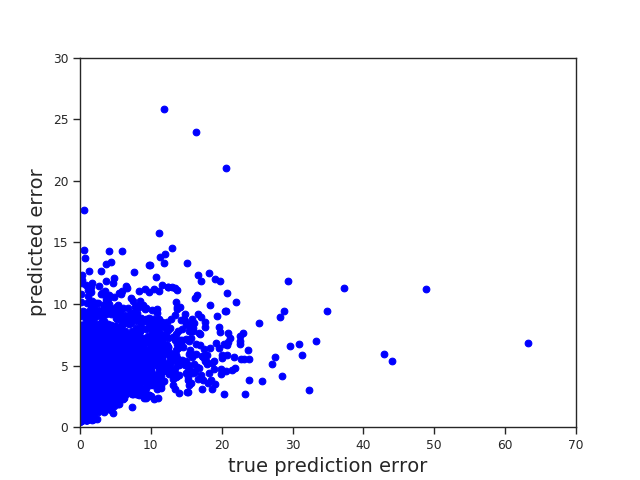}}
        \end{subfloat}
	\end{center}

  \caption{{\bf Model outputs for each image in the age estimation experiment.} The upper figures show the results of the joint formulation~\cite{NIPS2017_7141}, and the lower figures show the results of our method. The figures on the left side show the ground truth ($x$-axis) and predicted age ($y$-axis) of each sample in the test data.
The figures on the right side show the error of the age prediction ($x$-axis) and the prediction of the error ($y$-axis) of each sample.
}
\label{fig:age_scatter}
\end{figure}


\paragraph{Implementation Details}

We use a simple network composed of eight fully connected layers. We first feed the input into the layers, which have outputs of $32, 64, 128, 128, 64$, and $32$, respectively. After the last layer follows a pair of layers having outputs of $16$ and $1$; these predict the signal and uncertainty. Each layer except for the last two is followed by a ReLU. We divide the training data into 10 subsets for the training of each $M_j$.


\paragraph{Results}

Table~\ref{tbl:simulation_results} compares the values of the estimation metrics, and Figure~\ref{fig:simulation_pir} compares the PiR of the methods.
From Table~\ref{tbl:simulation_results}, it can be seen that our method has the best performance on the two metrics of uncertainty estimation. When $\Delta$ increases, which means that signal estimation becomes more difficult, the joint formulation becomes worse for all metrics. By contrast, our method seems to predict both the signal and uncertainty almost independently of the value of $\Delta$. Figure~\ref{fig:simulation_pir} compares the PiR($\eta$) of the frameworks, and it also shows that our method outperforms the others across a wide range of $\eta$.

Figure~\ref{fig:simulation_visualization} compares ourmethod to the joint formulation. When we look at the signal estimation, we see that the larger $\Delta$ is, the more unstable the signal estimation becomes in all the methods, and also that the task becomes difficult when $\Delta$ is large. By contrast, when we focus on the uncertainty estimation in Figure~\ref{fig:simulation_visualization}, the joint formulation is unstable with a large $\Delta$. We think that this instability is a result of being unable to balance the weight between the target and uncertainty estimation in the joint estimation. In the case of our method, the error estimation is stable and closer to the ideal mean error given by Eq.~\ref{eq:simulation_variance}. The error estimation is larger when $x$ is larger, but this is because the target estimation is inappropriate for this noisy area. 


\subsection{Age Estimation}
\label{subsec:age_estimation}

In this experiment, we predict ages and their uncertainty from facial images.

\paragraph{Datasets}

We use the UTKFace dataset~\cite{zhifei2017cvpr} for the age estimation. The UTKFace dataset is a large-scale facial dataset consisting of more than 20,000 data. It is composed of cropped images of the faces of people ranging in age from 0 to 116 years. They used DEX~\cite{Rothe2016IJCV} algorithm to guess the ground truth of the ages, and then checked it manually. We randomly choose 80\% of the data for training, 10\% for validation, and the remaining 10\% for testing. For the inputs of the networks, we resize the images into a resolution of $64 \times 64$.

\begin{table}[!t]
\small
  \begin{center}
  \caption{{\bf Depth estimation results on NYU Depth Dataset V2}. All metrics were calculated for each image; the table shows averages over all images.}
  \label{tbl:depth_results}
  \begin{tabular}{|c||c||c|c|} \hline
    loss & Target & eRMSE & PiR($0.5$) \\
    & RMSE ($\downarrow$) & ($\downarrow$) & ($\uparrow$) \\ \hline \hline
	$L1$ & 0.529 & - & - \\ \hline
    JF~\cite{NIPS2017_7141}  & 0.531 & 0.482  & 0.364 \\ \hline
    our method ($\lambda = 0.05$)  & 0.522 & {\bf 0.399} & {\bf 0.401} \\ \hline 
  \end{tabular}
  \end{center}
\end{table}


\begin{figure}[!t]
\begin{center}
    \begin{subfloat}{
          \includegraphics[clip, width=0.47\columnwidth]{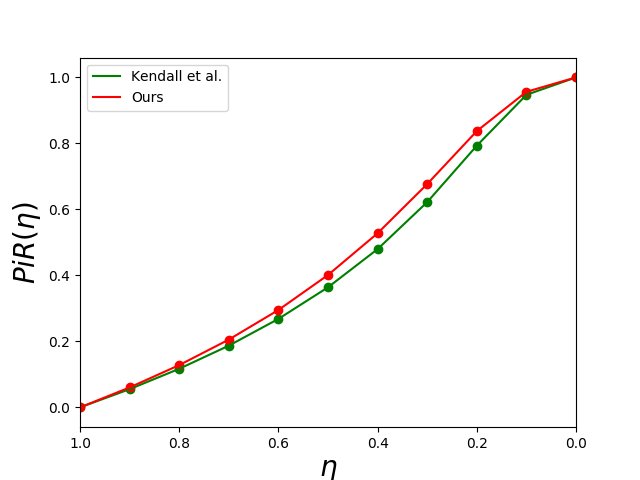}}
        \end{subfloat}
\begin{subfloat}{
          \includegraphics[clip, width=0.47\columnwidth]{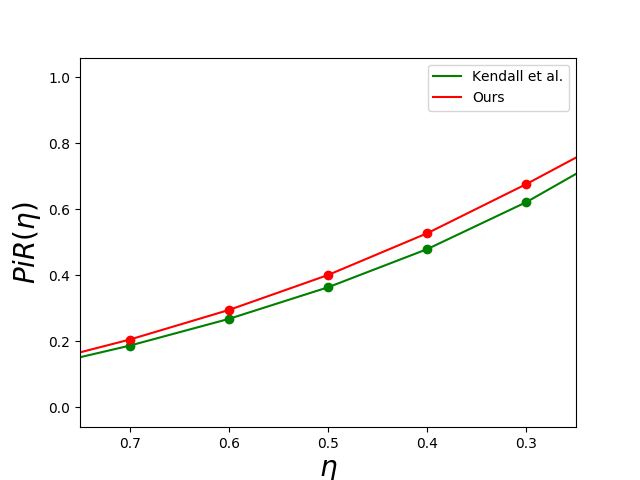}}
        \end{subfloat}
    \caption{{\bf PiR of the depth estimation.} 
The left image shows the value of PiR($\eta$), where $\eta$ decreases by $0.1$ from $1$ to $0$. The enlarged detail on the right shows the region where $\eta$ decreases from $0.7$ to $0.3$. The green and red curves are the results of the joint formulation~\cite{NIPS2017_7141} and our method, respectively. 
}
    \label{fig:depth_pir}
  \end{center}
\end{figure}

\paragraph{Implementation Details}

We use ResNet50~\cite{he2016deep}, which is pre-trained on ImageNet~\cite{deng2009imagenet}. This pre-trained model is provided on PyTorch. We change the last layer of the provided model to regress two values: the age and the certainty of the estimation. During this experiment, we divide the training data into 20 subsets for the training of each $M_j$.


\begin{figure*}[!t]
	\begin{center}
  \begin{subfloat}[Input]{
          \includegraphics[clip, width=0.23\columnwidth]{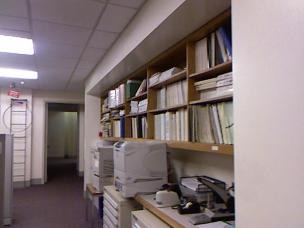}}
    \end{subfloat}
\begin{subfloat}[GT]{
          \includegraphics[clip, width=0.23\columnwidth]{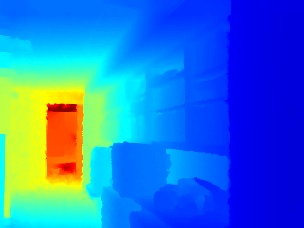}}
    \end{subfloat}
    \begin{subfloat}[Pred. (JF)]{
          \includegraphics[clip, width=0.23\columnwidth]{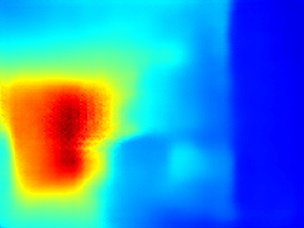}}
        \end{subfloat}
        \begin{subfloat}[Error (JF)]{
          \includegraphics[clip, width=0.23\columnwidth]{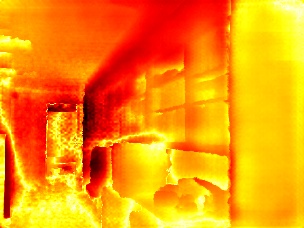}}
        \end{subfloat}
        \begin{subfloat}[Unc. (JF)]{
          \includegraphics[clip, width=0.23\columnwidth]{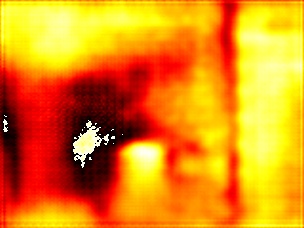}}
        \end{subfloat}
\begin{subfloat}[Pred. (ours)]{
          \includegraphics[clip, width=0.23\columnwidth]{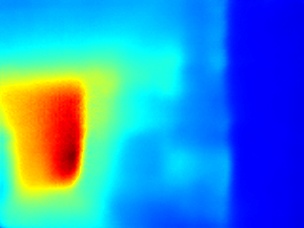}}
    \end{subfloat}
\begin{subfloat}[Error (ours)]{
          \includegraphics[clip, width=0.23\columnwidth]{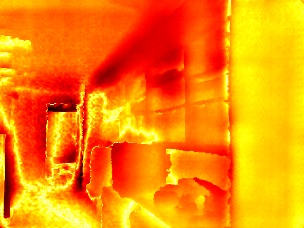}}
        \end{subfloat}
\begin{subfloat}[Unc. (ours)]{
          \includegraphics[clip, width=0.23\columnwidth]{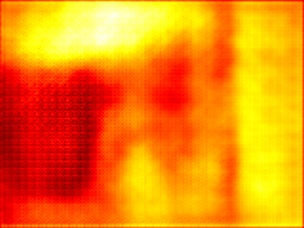}}
        \end{subfloat}
        \vspace{2mm}
\begin{subfloat}[Input]{
          \includegraphics[clip, width=0.23\columnwidth]{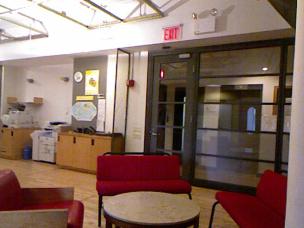}}
    \end{subfloat}
\begin{subfloat}[GT]{
          \includegraphics[clip, width=0.23\columnwidth]{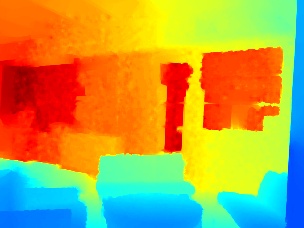}}
    \end{subfloat}
    \begin{subfloat}[Pred. (JF)]{
          \includegraphics[clip, width=0.23\columnwidth]{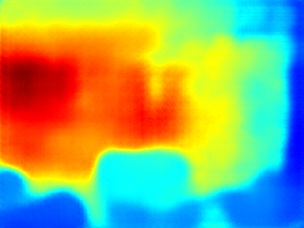}}
        \end{subfloat}
        \begin{subfloat}[Error (JF)]{
          \includegraphics[clip, width=0.23\columnwidth]{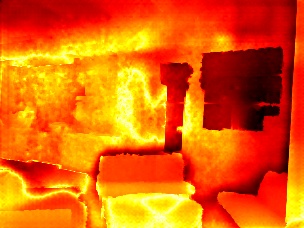}}
        \end{subfloat}
        \begin{subfloat}[Unc. (JF)]{
          \includegraphics[clip, width=0.23\columnwidth]{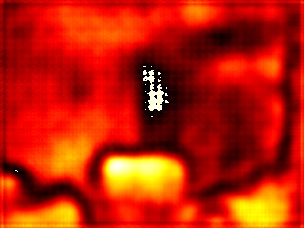}}
        \end{subfloat}
\begin{subfloat}[Pred. (ours)]{
          \includegraphics[clip, width=0.23\columnwidth]{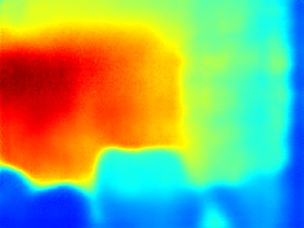}}
    \end{subfloat}
\begin{subfloat}[Error (ours)]{
          \includegraphics[clip, width=0.23\columnwidth]{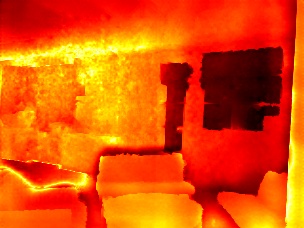}}
        \end{subfloat}
\begin{subfloat}[Unc. (ours)]{
          \includegraphics[clip, width=0.23\columnwidth]{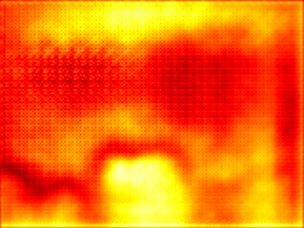}}
        \end{subfloat}
        \vspace{2mm}
  \begin{subfloat}[Input]{
          \includegraphics[clip, width=0.23\columnwidth]{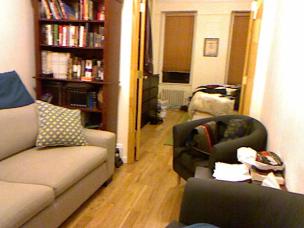}}
    \end{subfloat}
\begin{subfloat}[GT]{
          \includegraphics[clip, width=0.23\columnwidth]{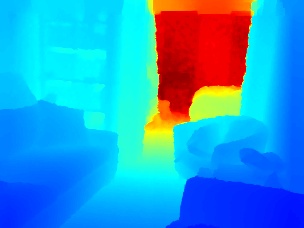}}
    \end{subfloat}
    \begin{subfloat}[Pred. (JF)]{
          \includegraphics[clip, width=0.23\columnwidth]{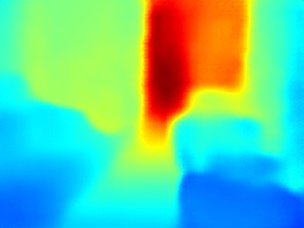}}
        \end{subfloat}
        \begin{subfloat}[Error (JF)]{
          \includegraphics[clip, width=0.23\columnwidth]{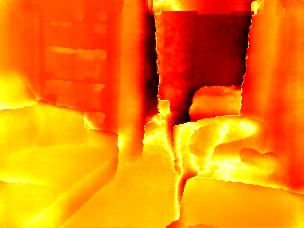}}
        \end{subfloat}
        \begin{subfloat}[Unc. (JF)]{
          \includegraphics[clip, width=0.23\columnwidth]{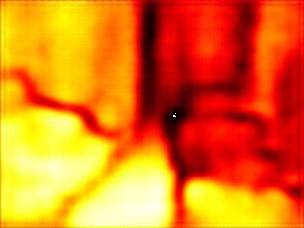}}
        \end{subfloat}
\begin{subfloat}[Pred. (ours)]{
          \includegraphics[clip, width=0.23\columnwidth]{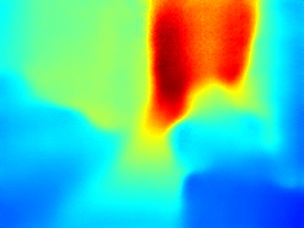}}
    \end{subfloat}
\begin{subfloat}[Error (ours)]{
          \includegraphics[clip, width=0.23\columnwidth]{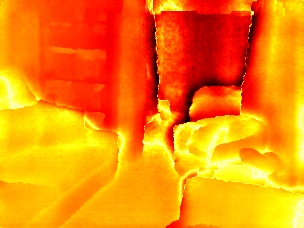}}
        \end{subfloat}
\begin{subfloat}[Unc. (ours)]{
          \includegraphics[clip, width=0.23\columnwidth]{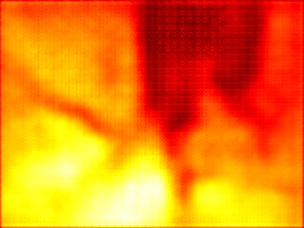}}
        \end{subfloat}
	\end{center}

	\caption{{\bf Examples of depth estimation.}
From left to right: input RGB images (Input), ground-truth (GT) depth maps, joint formulation (JF)~\cite{NIPS2017_7141} predicted depth map (Pred.), JF absolute errors (Error), JF expected uncertainty map (Unc.), and the three corresponding images from our method. In the ground truth and predicted depth maps, the black or red regions have large depth values, whereas the blue or green regions contain small depth values. In the error and uncertainty maps, the black or red regions have larger values, whereas the white or yellow regions contain smaller values.
}
	\label{fig:depth_examples}
\end{figure*}


\begin{figure}[!t]
\begin{center}
    \includegraphics[width=0.47\columnwidth]{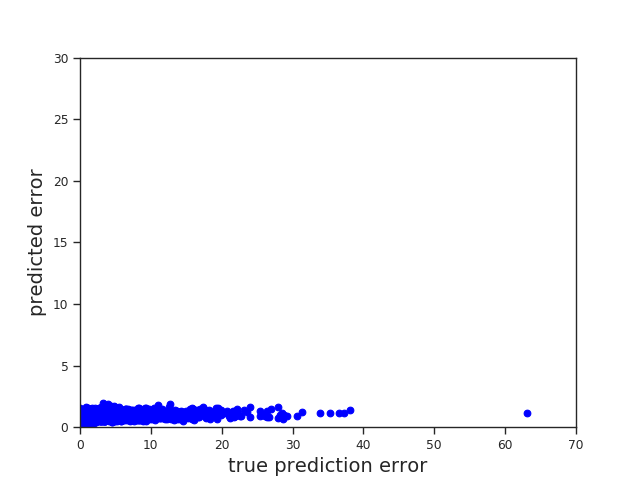}
    \caption{{\bf Plot of model outputs for each image in the age estimation}. The error of the age prediction of each sample is shown on the $x$-axis, and the prediction of the error on the $y$-axis. The model is trained with the separate formulation without virtual residuals.}
    \label{fig:age_estimation_ablation}
  \end{center}
\end{figure}


\paragraph{Results}

Table~\ref{tbl:age_results} compares the age estimation in the two frameworks of uncertainty estimation, and in a simple $L1$ loss estimation without uncertainty estimation. Figure~\ref{fig:age_pir} compares the PiR($\eta$) of the frameworks. They show that our method outperforms the existing methods in terms of eRMSE and PiR.

To analyze the differences, we plot two sets of pairs in Figure~\ref{fig:age_scatter}: ground truth and predicted ages, and actual and estimated prediction errors. A comparison of the two figures for the age estimation shows that the models estimate the age similarly; both approximately follow the line $y = x$. By contrast, the figures for the error estimation are different. The existing method, JF, uniformly predicts the error to be small, even when the real error is large, whereas our method predicts a larger error in response to a larger real estimation error. The models (including ours) always estimate the error as less than 30 years. This is probably because the number of images that have such large errors is few.

Table~\ref{tbl:age_examples} shows two examples of age estimation. The existing method, JF, tends to underestimate the error. When the error is larger, this tendency is clearer. The underestimation of large errors is especially mitigated by our method.


\subsection{Depth Estimation}
\label{subsec:depth_estimation}

Finally, we compare the models based on a depth estimation task using RGB images. This task is inherently difficult~\cite{Ma2018SparseToDense}.

\paragraph{Datasets}

We use a popular RGB-based depth prediction task, NYU Depth V2~\cite{Silberman:ECCV12}.
This dataset consists of RGB and depth images collected from 464 different indoor scenes using Microsoft Kinect. We use the official split of the data: 249 scenes are used for training, and the remaining 215 are used for testing, as in~\cite{Ma2018SparseToDense,laina2016deeper,eigen2014depth}. 


\paragraph{Implementation Details}

We implement our proposed method and the existing ones at the top of the official sparse-to-dense~\cite{Ma2018SparseToDense} PyTorch implementation\footnote{\url{github.com/fangchangma/sparse-to-dense.pytorch}}. Note that we do not use any sparse depth sample as an input. We increase the number of epochs to 30, because it is difficult in this task for the models to achieve convergence. We implement the network layers in the same way as ~\cite{asai}, using ResNet~\cite{he2016deep} and UpProj~\cite{laina2016deeper}.
(For more details, please refer to ~\cite{asai}.)
During this difficult task, we divide the training data into 30 subsets for the training of each $M_j$.


\paragraph{Results}

In this experiment, the depth and uncertainty are estimated per pixel, and thus we calculate all metrics per image. Table~\ref{tbl:depth_results} compares the depth estimation of the two uncertainty estimation frameworks and a simple $L1$ loss estimation. Figure~\ref{fig:depth_pir} compares the PiR($\eta$) of the two frameworks. The values of eRMSE and PiR($\eta$) in our framework exceed those in JF. This task is difficult for neural networks to predict ~\cite{Ma2018SparseToDense}, and thus the epistemic uncertainty of the networks is larger. In such a case, quantifying the error from the aleatoric uncertainty is much more difficult, resulting in smallness of the difference of performance between the methods.

Figure~\ref{fig:depth_examples} shows examples of the estimations. In the JF estimation, there are areas where the error estimation is inappropriately small, and there are some white areas in where the errors are large. Additionally, JF is less sensitive to the wall, where the errors tend to be larger. Our method avoids these phenomena.


\subsection{The benefit of the virtual residuals}

Lastly, we argue for the benefit of introducing virtual residuals to optimize $w_i$. 

We conduct experiments without using virtual residuals -- we substitute the value $r_i$ directly for $\tilde{r_i}$ in Eq.~\ref{eq:ours_laplace} instead of virtual residuals. We treat $\tilde{r}_i$ as a constant in a minibatch and do not calculate the gradient because we want to balance the weight of the two terms, $L_t$ and $L_u$, and focus on the ablation of the virtual residuals.

Figure~\ref{fig:age_estimation_ablation} shows the prediction error and its estimation in the age estimation case. Comparing this with Figure~\ref{fig:age_scatter}, we can see that the separable formulation without virtual residuals estimates the error to be much smaller than the joint formulation. We consider this to be robustness of the target estimation in Eq.~\ref{eq:kendall_laplace}. The partial derivative of Eq.~\ref{eq:kendall_laplace} with respect to $\theta$ is
\begin{equation}
  \frac{\partial\mathcal{L}}{\partial\theta} = \frac{1}{N}\sum_{i=1}^N \left\{ \exp(w_i)\frac{\partial r_i}{\partial \theta} + \exp(w_i) r_i \frac{\partial w_i}{\partial \theta} - \frac{\partial w_i}{\partial \theta} \right\}.
  \label{eq:kendall_laplace_derivative}
\end{equation}
When given noisy samples with large aleatoric uncertainties, $w_i$ becomes small in the training. Therefore the first term of Eq.~\ref{eq:kendall_laplace_derivative} also becomes small. It implies that the signal estimation by Eq.~\ref{eq:kendall_laplace} is robust. (This effect is also mentioned in ~\cite{NIPS2017_7141} as attenuating loss.) However, this effect is not sufficient as we can see in the experiments. The virtual residuals avoid underestimating.


\section{Conclusion}
\label{sec:conclusion}

In this study, we presented a framework for error quantification based on  heteroscedastic aleatoric uncertainty estimation in regression problems of deep learning. Existing methods cannot estimate the error accurately because of overfitting in the target estimation. We mitigated the effect of overfitting with virtual residuals and avoided underestimating the estimation errors. In the experiments, we evaluated our method on three different tasks, and showed that it estimated both targets and errors, outperforming the existing method on many metrics.


\section*{Acknowledgment}
This work was partially supported by JST JPMJCR19F4 and JSPS 18H03254.

\bibliographystyle{IEEEtran}


\end{document}